\documentclass{vgtc}                          

\graphicspath{{figs/}{tabs/}{images/}{./}} 

\usepackage{times}                     

\usepackage{tabu}                      
\usepackage{booktabs}                  
\usepackage{lipsum}                    
\usepackage{mwe}                       

\usepackage{mathptmx}                  
\usepackage{amsmath} 
\usepackage{amssymb}
\usepackage{tabularx}

\newcommand{\eg}{\textit{e.g.}\xspace}

\newcommand{\etc}{\textit{etc.}\xspace}

\onlineid{0}

\vgtccategory{Research}

\vgtcinsertpkg

\title{BrainDreamer: Reasoning-Coherent and Controllable Image Generation from EEG Brain Signals via Language Guidance}

\author{Ling Wang$^{\dagger}$\\ %
        \parbox{1.8in}{\scriptsize \centering AI Thrust, HKUST(GZ)\\lwang851@connect.hkust-gz.edu.cn} %
\and Chen Wu$^{\dagger}$\\ %
     \parbox{1.8in}{\scriptsize \centering USTC \\ wuchen5X@mail.ustc.edu.cn} %
\and Lin Wang\thanks{Corresponding author. $^\dagger$ Equal contribution.}\\ %
     \parbox{1.8in}{\scriptsize \centering AI/CMA Thrust, HKUST(GZ) \\ Dept. of CSE, HKUST \\ linwang@ust.hk}}
\teaser{
  \centering
  \includegraphics[width=\linewidth]{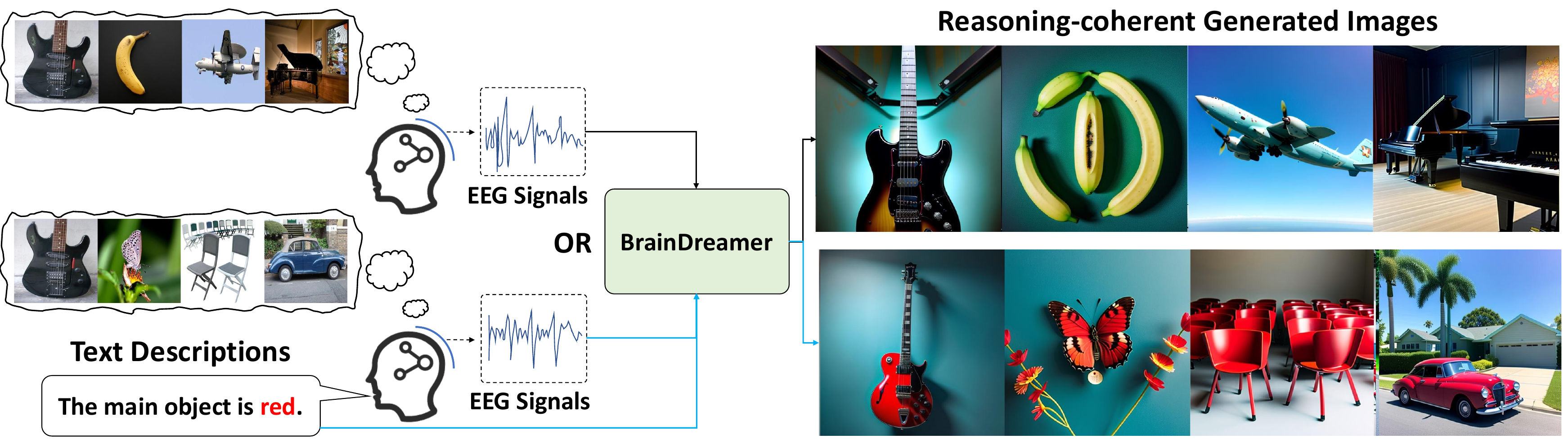}
  \vspace{-15pt}
  \caption{\textbf{An overview of our BrainDreamer}, a novel language-guided EEG-to-Image generation framework. It can generate high-quality reasoning-coherent images from EEG brain signals (black arrowed line). In addition, it can take additional textual (\eg, color, position, \etc.) descriptions as guidance to achieve controllable image generation (blue arrowed line).}
  \label{fig:teaser}
}

\abstract{
\emph{Can we directly visualize what we imagine in our brain together with what we describe?} The inherent nature of human perception reveals that, when we think, our body can combine language description and build a vivid picture in our brain. Intuitively, generative models should also hold such versatility. In this paper, we introduce \textbf{BrainDreamer}, a novel end-to-end language-guided generative framework that can mimic human reasoning and generate high-quality images from electroencephalogram (EEG) brain signals. Our method is superior in its capacity to eliminate the noise introduced by non-invasive EEG data acquisition and meanwhile achieve a more precise mapping between the EEG and image modality, thus leading to significantly better-generated images.
Specifically, BrainDreamer consists of two key learning stages: \textbf{1)} modality alignment and \textbf{2)} image generation. In the alignment stage, we propose a novel \textit{mask-based triple contrastive learning strategy} to effectively align EEG, text, and image embeddings to learn a unified representation. 
In the generation stage, we inject the EEG embeddings into the pre-trained Stable Diffusion model by designing a learnable EEG adapter to generate high-quality reasoning-coherent images. Moreover, BrainDreamer can accept textual descriptions (\eg, color, position, \etc) to achieve \textit{controllable} image generation. Extensive experiments show that our method significantly outperforms prior arts in terms of generating quality and quantitative performance.
} 

\keywords{EEG, Image generation, Diffusion model, Perception and cognition}

\begin{document}



\firstsection{Introduction}
\maketitle
\begin{figure*}[t!]
    \centering
    \includegraphics[width=0.74\textwidth]{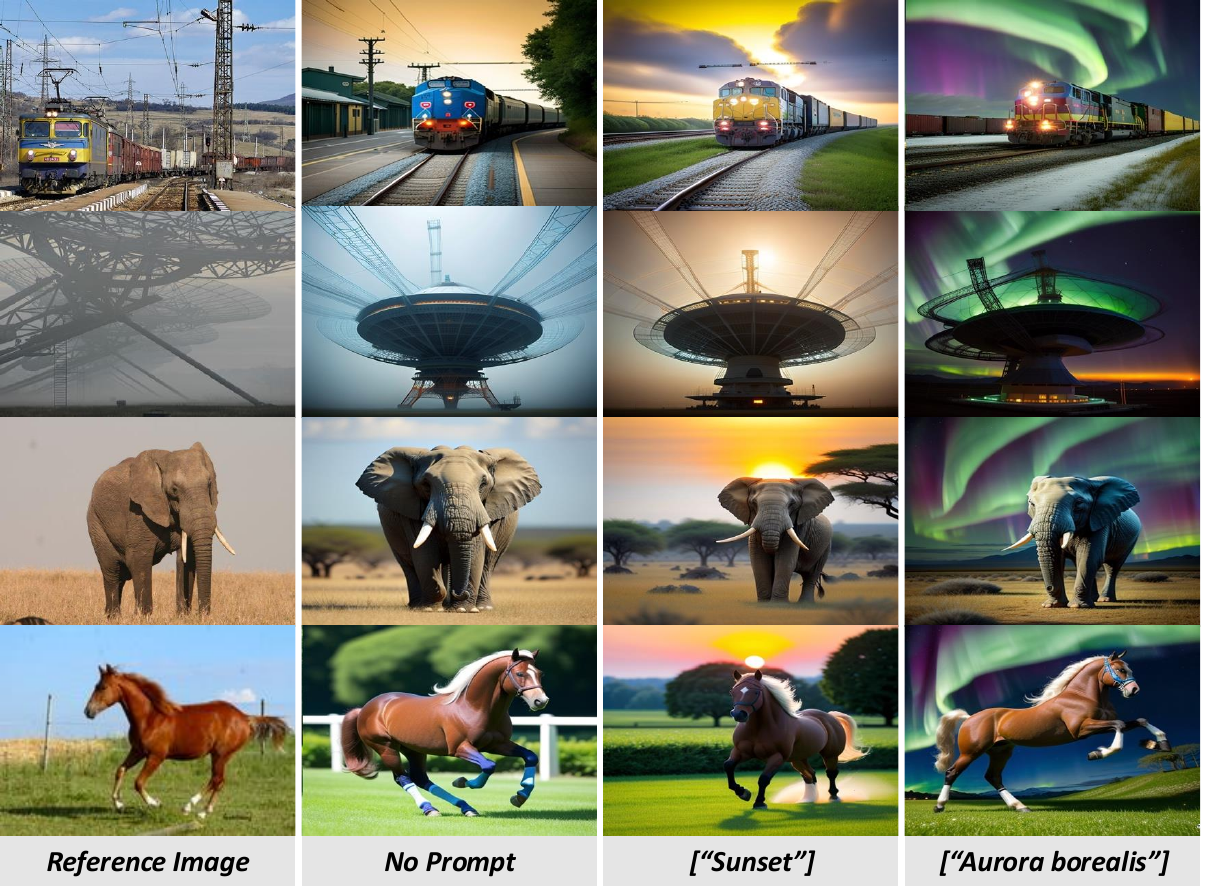}
    \vspace{-8pt}
    \caption{Diverse generation and creation results of our BrainDreamer with text guidance. BrainDreamer can achieve high-quality, reasoning-coherent, and controllable image generation from different textual descriptions, such as \textsf{[``Sunset"]}, \textsf{[``Aurora borealis"]}.}
     \vspace{-10pt}
    \label{fig:res_text}
\end{figure*}

Human visual perception has long remained enigmatic. This has attracted neuroscientists and AI researchers to investigate the mechanisms of human vision. With the advancement of deep learning, generative models have reached a point where they can directly generate similar observed images from brain signals~\cite{bai2023dreamdiffusion,kavasidis2017brain2image,singh2023eeg2image,singh2024learning,zeng2023dm,lu2023minddiffuser,chen2023seeing,kupershmidt2022penny,zeng2023controllable}. However, directly generate images from EEG signals may not align with human cognitive processes or human-computer interaction paradigms, as such direct generation could result in significant inaccuracies or omissions in detail. When we imagine a scene, it often undergoes iterative refinement from coarse to fine details and can be supplemented or corrected based on different descriptions. Therefore, in addition to possessing the ability to generate observed images from brain signals directly, the generative models should also be capable of accepting textual descriptions to assist in image generation. This capability has profound implications for virtual reality (VR) content creation, where content creation tools that incorporate brain signals and textual inputs could offer a more intuitive and human-centered approach. For instance, a user imagining a rough scene in a VR environment could iteratively refine it through mental imagery and further enhance it with verbal descriptions. By incorporating textual input to refine fine-grained details or correct errors in the generated imagery, these models could become more useful for VR content creation. Additionally, such advancements open new possibilities in fields like personalized entertainment and art creation.
In this paper, we investigate how to generate observed images from EEG brain signals in a manner that aligns more closely with human visual perception. 
EEG is a non-invasive technique that stands as the most commonly used method for capturing the electrophysiological dynamics of the brain~\cite{sakkalis2011applied,tian2024says}. EEG data typically refers to time-series electrophysiological signals that are recorded using electrodes positioned on the human scalp~\cite{spampinato2017deep,cheng2024enhancing}. These recordings are often conducted while subjects are presented with stimuli, \eg, watching some images within a specific time frame. Exploring the connection between EEG signals and brain activity is a highly meaningful endeavor~\cite{bai2023dreamdiffusion,singh2024learning}. It enables the recording of our momentary thoughts, and more significantly, it holds the potential to assist in the treatment of individuals with conditions such as cognitive impairments.

Recently, there have been some explorations for generating images from the EEG data~\cite{bai2023dreamdiffusion,kavasidis2017brain2image,singh2023eeg2image,singh2024learning,zeng2023dm}. While they demonstrate promising outcomes, they encounter various limitations. Firstly, EEG data are captured non-invasively and thus are inherently noisy. Some methods~\cite{kavasidis2017brain2image,singh2023eeg2image,singh2024learning,zeng2023dm} disregard the noise, leading to lower quality in the generated images. Secondly, several methods~\cite{bai2023dreamdiffusion,singh2024learning} attempt to utilize single-label information to guide the network in learning deep features from EEG data to achieve alignment with EEG-Image pairs. However, in most cases, single-label information is insufficient to depict complex image scenarios, resulting in inaccurate instances in the generated images. In a nutshell, \textit{these methods either ignore the noise introduced by non-invasive EEG acquisition or fail to achieve a precise mapping between EEG and image modality, resulting in poor-quality generated images}. Moreover, previous works~\cite{bai2023dreamdiffusion,kavasidis2017brain2image,singh2023eeg2image,singh2024learning,zeng2023dm} solely focus on generating observed images from EEG data, \textit{disregarding the supplementary information such as textual descriptions, which does not conform to human visual perception}. 

To address the aforementioned challenges, we propose a novel framework, called \textbf{BrainDreamer}, as depicted in Fig.~\ref{fig:teaser}. It can mimic human reasoning and generate high-quality images from EEG brain signals. It can also accept supplementary information such as textual descriptions, which better conform to human visual perception. It enhances user engagement by aligning VR content more closely with an individual’s mental imagery and preferences, fostering a more intuitive and responsive interaction. This methodology opens new possibilities for creating VR content that is not only visually compelling but also more attuned to the nuances of human perception, making VR experiences more natural and human-centered. Specifically, our method consists of two key learning stages: modality alignment and image generation. 
\begin{figure*}[t!]
    \centering
    \includegraphics[width=.84\textwidth]{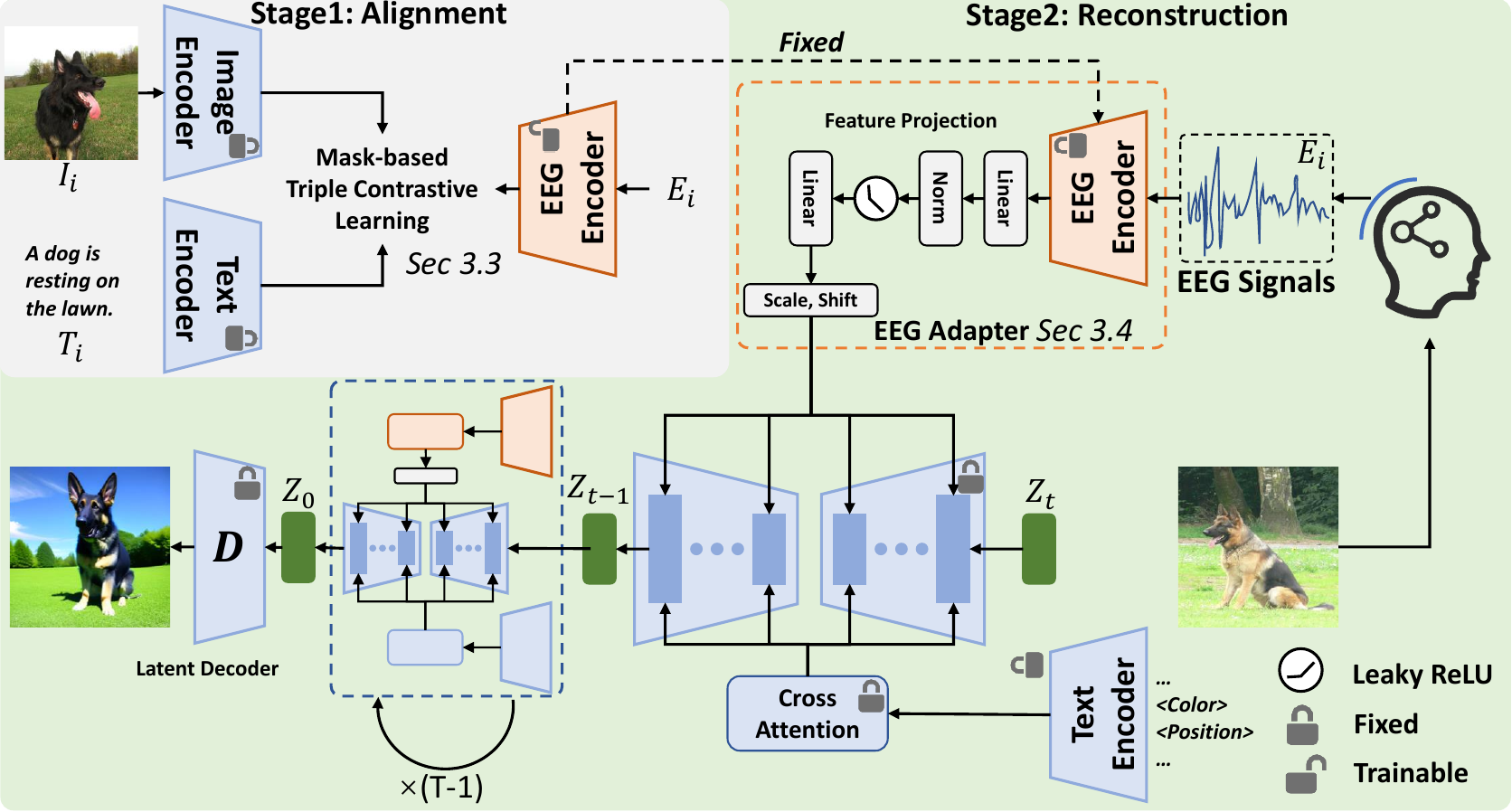}
    \vspace{-5pt}
    \caption{\textbf{Overview of our BrainDreamer.} After aligning the EEG signals, images, and text using a mask-based triple contrastive learning strategy, we design an EEG adapter based on the trained EEG encoder. The EEG adapter employs the FiLM to modulate EEG embeddings. Then, the EEG and text embeddings are fed into pre-trained Stable Diffusion to generate reasoning-coherent images.}
    \vspace{-10pt}
    \label{fig:framework}
\end{figure*}
In the modality alignment stage, we leverage Contrastive Language-Image Pre-training (CLIP)~\cite{clip} to assist in aligning EEG, text, and image embeddings. CLIP learns a multi-modal embedding space shared by the text and image feature and contains a wide range of visual concepts. We design a novel mask-based triple contrastive learning strategy to map EEG embeddings into the CLIP embeddings space (Sec.~\ref{sec:cl}). Prior approaches~\cite{bai2023dreamdiffusion,kavasidis2017brain2image,singh2023eeg2image,singh2024learning,zeng2023dm} predominantly emphasized the alignment of EEG data with images, neglecting the significant \textit{semantic} information that text can offer. Introducing text information to supervise EEG embeddings can make them more flexible and controllable. Also, we employ the masked modeling~\cite{mae,flip} on the image and EEG data. During the training, random masks are applied to both the image and EEG data to discard certain information. Such an approach not only enhances the robustness of features (\eg, reducing noise interference and alleviating inter-individual differences in EEG data) but also reduces the training cost~\cite{flip}. 

In the image generation stage, we design an EEG adapter to inject EEG embeddings into the pre-trained Stable Diffusion~\cite{sd}(Sec.~\ref{sec:eeg}). Specifically, the EEG adapter consists of a frozen EEG encoder and a feature projection module that is dedicated to reducing the domain gap between EEG embeddings and CLIP embeddings. Subsequently, the output of the feature projection module is considered as scale and shift factors. Then, the EEG adapter employs a Feature-wise Linear Modulation (FiLM)~\cite{film} to inject EEG embeddings into the pre-trained Stable Diffusion to generate images based on the scale and shift factors. This offers a lower computational overhead compared to commonly used cross-attention methods. Particularly, BrainDreamer is capable of incorporating additional textual descriptions, such as color and position information, to assist in the generation of images. This way, our BrainDreamer can generate high-quality, controllable, and reasoning-coherent images from EEG signals, as demonstrated in Figs.~\ref{fig:teaser}, and \ref{fig:res}. For example, when we input text \textsf{[``the main object is red'']} or \textsf{[``on the beach, seaside'']}, our method can make the main subject color of the generated image red or the background of the generated image is a beach, located by the seaside. Our approach allows for a more interactive, adaptive and fluid content creation process, where users can iteratively refine their contextually rich visual outputs based on feedback and preferences. 

%
In summary, our main contributions are three-fold: (\textbf{I}) We propose BrainDreamer that can mimic human reasoning and generate high-quality images from EEG brain signals. Moreover, our BrainDreamer can also incorporate additional textual descriptions to assist in generating reasoning-coherent images from EEG brain signals. 
(\textbf{II}) We propose a mask-based triple contrastive learning strategy to effectively align EEG, text, and image embeddings. In the generation stage, we propose an EEG adapter to inject EEG embeddings into the pre-trained Stable Diffusion model to generate high-quality reasoning-coherent images. (\textbf{III}) We intensively evaluate BrainDreamer, and the results show that our BrainDreamer significantly outperforms prior arts in terms of generation quality and quantitative performance.

\section{Related Work}
\label{sec:related}
\noindent \textbf{Image Generation from EEG Signals.}
The rapid development of learning-based methods has made it possible to extract meaningful representations from brain signals, such as EEG. Brain2Image~\cite{kavasidis2017brain2image} is the first to generate observed images of ImageNet from EEG features. EEGStyleGAN-ADA~\cite{singh2024learning} improves image synthesis from EEG signals by leveraging learned EEG representations in contrastive settings. As diffusion models have shown remarkable achievements in image generation, DM-RE2I~\cite{zeng2023dm} attempts to incorporate EEG features into the diffusion model by adding them to the time step. The current state-of-the-art method, DreamDiffusion~\cite{bai2023dreamdiffusion}, utilizes the pre-trained Stable Diffusion model to generate high-quality images from EEG signals. It employs large-scale additional datasets (approximately 120K samples) for self-supervised pre-training and aligns EEG embeddings with CLIP image embeddings to enhance the EEG encoder's representation capability of EEG features. However, it requires fine-tuning all parameters of the diffusion models, which brings about significant training costs. Moreover, it merely focuses on aligning EEG data with images, overlooking the significant semantic information that text can offer. Furthermore, these methods can only directly generate images from EEG brain signals, without the ability to make adjustments to the generated images based on supplementary information (\eg, textual descriptions). This does not align with human visual perception. 

\noindent \textbf{Pre-training Vision-Language Models (VLMs).}
The recent advancements in vision and language models have significantly advanced the integration and comprehension of both image and language modalities. CLIP~\cite{clip} achieves the alignment of image and text features by leveraging contrastive learning on large-scale image-text datasets. It maps images and text into a unified embedding space and demonstrates excellent zero-shot capabilities across various visual and language tasks. BLIP~\cite{blip} refines pre-trained methods by enhancing the quality of image and text data by the image captioner. Building upon this,  BLIP2~\cite{blip2} leverages pre-trained VLMs to enhance the model's multimodal understanding capability while significantly reducing training costs. InstructBLIP~\cite{instructblip} allows us to perform various tailored vision tasks through instruction tuning. Furthermore, state-of-the-art VLMs such as GPT-4~\cite{gpt4} and Gemini~\cite{team2023gemini}, they possess powerful representational capabilities and can handle a wide range of multimodal tasks. In this work, We leverage CLIP and propose a mask-based triple contrastive loss (see Eq.~\ref{eq:loss}) to assist us in aligning EEG, text, and image embeddings. In contrast to previous works, we introduce text supervision to assist in achieving better embedding alignment.

\noindent \textbf{Diffusion Models.}
In recent years, we have witnessed the incredible results of diffusion probabilistic models~\cite{DDPM} in controllable image generation, particularly in text-to-image generation. GLIDE~\cite{nichol2021glide} adopts a cascaded text-guided diffusion architecture to support both image generation and editing. Imagen~\cite{imagen} encodes text to embeddings using language models as the condition into diffusion model, achieving high fidelity of the generated image. Stable Diffusion~\cite{sd} moves the execution of the diffusion process to the latent space instead of the original pixel space, which significantly reduces the computation cost. In addition, It extends the types of control conditions beyond just text, allowing for the inclusion of depth maps, semantic segmentation maps, and more. Subsequently, a series of fine-tuning works based on Stable Diffusion emerged. ControlNet~\cite{control} controls diffusion model with task-specific conditions by fine-tuning a “trainable copy” of any off-the-shelf diffusion model. UniControl~\cite{unicontrol} incorporates task instruction, which enables a single model to handle multiple condition-to-image tasks. To achieve efficient fine-tuning, IP-Adapter~\cite{ipadapter} employs a decoupled cross-attention strategy to reduce the trainable parameters to 22M.  We design an EEG adapter to inject EEG embeddings into the pre-trained Stable Diffusion model, enabling image generation with the integration of text descriptions.

\begin{figure*}[t!]
    \centering
    \includegraphics[width=.78\textwidth]{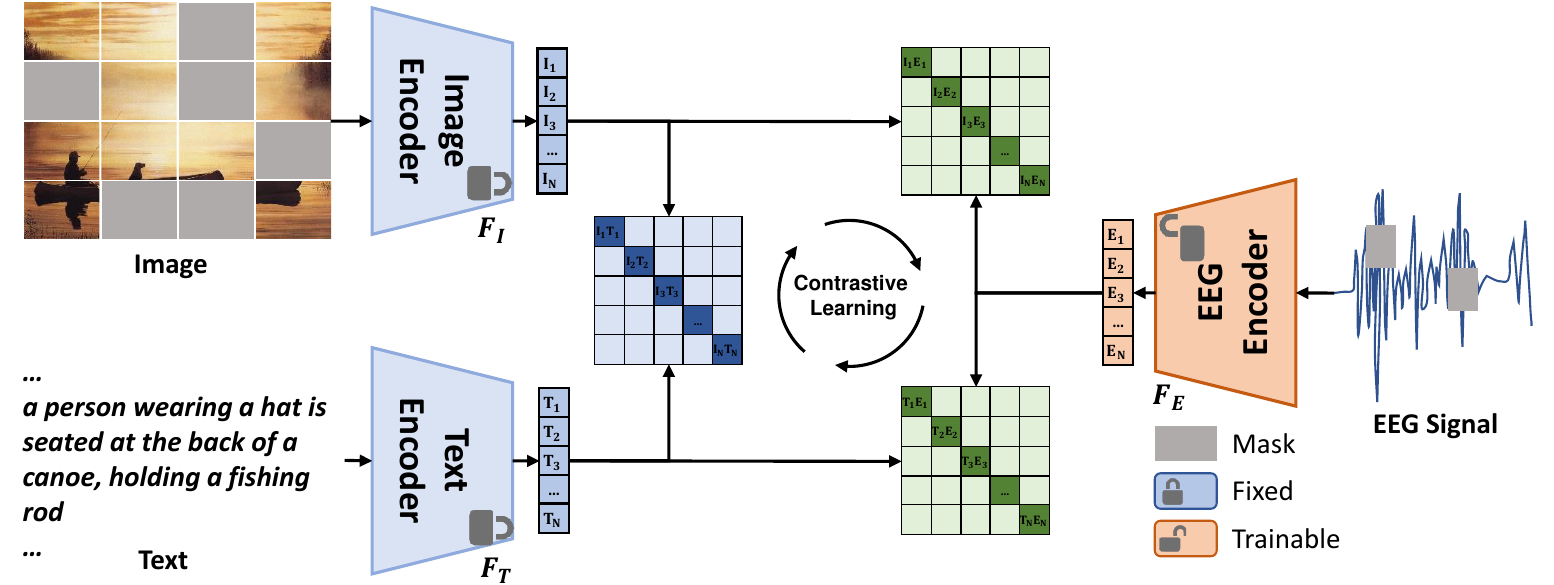}
    \caption{\textbf{Mask-based triple contrastive learning.} We leverage CLIP's image and text encoders to assist in training the EEG encoder. Also, during training, random masks are applied to both the image and EEG data to enhance feature robustness and reduce training costs.}
    \label{fig:stage1}
    \vspace{-10pt}
\end{figure*}

\section{The Proposed Method}
\label{sec:methods}
\subsection{Prelimiaries and Background}
Diffusion models are one type of generative model, consisting of a forward (\textit{a.k.a}, the forward process) and a backward process. In the forward process, Gaussian noise with variance $\beta_{t}\in(0,1)$ at time $\boldsymbol{t}$ is added to the input $\boldsymbol{x_{0}}$ for producing the noisy input. At each step $t\in\{0,\ldots,T\}$, the intermediate sample $\boldsymbol{x_{t}}$ is computed as:
\begin{equation}
    \boldsymbol{x_t}=\sqrt{\bar{\alpha_t}}\boldsymbol{x_0}+\sqrt{1-\bar{\alpha_t}}\boldsymbol{\epsilon_t},
\end{equation}
where $\boldsymbol{\epsilon_{t}}\sim\mathcal{N}(0,1)$ is the Gaussian noise at step $t$, $\alpha_{t}=1-\beta_{t}$ and $\bar{\alpha}_{t}=\prod_{s=1}^{t}\alpha_{s}$. When $t$ is large enough, the input $x_{t}$ is nearly a standard Gaussian distribution. 

A network $\boldsymbol{\epsilon_{\theta}}$ is learned by predicting the noise $\boldsymbol{\epsilon_{t}}$ conditioned on $\boldsymbol{c}$ (\eg, text prompts) at a randomly picked time-step $t$. The optimization of the diffusion model is defined as follows:
\begin{equation}
    \mathcal{L}_{dm}=\mathbb{E}_{x_{0},c,t,\epsilon}[||\boldsymbol{\epsilon_{t}}-\boldsymbol{\epsilon_{\theta}}(\boldsymbol{x_{t}}=\sqrt{\bar{\alpha}_{t}}\boldsymbol{x_{0}}+\sqrt{1-\bar{\alpha}_{t}}\boldsymbol{\epsilon_{t}},\boldsymbol{c},t)||^{2}],
\label{eq:L_dm}
\end{equation}
where $||\cdot||^{2}$ is the $\mathcal{L}_{2}$ loss.

We implement our method based on the pretrained Stable Diffusion~\cite{sd}, which is a latent text-to-image diffusion model. This model relies on an autoencoder that converts an image $\boldsymbol{x}$ into a latent $\boldsymbol{z}$ with encoder ${\mathcal{E}}$ to achieve better efficiency and stabilized training. Then generates it with decoder ${\mathcal{D}}$ after completing the forward process. In the sampling stage, Stable Diffusion also randomly drops out $\boldsymbol{c}$ to reduce reliance on the conditions. In other words, the predicted noise is calculated based on the prediction of both the conditional model $\boldsymbol{\epsilon}_\theta(\boldsymbol{z}_{t},\boldsymbol{c},t)$ and unconditional model $\boldsymbol{\epsilon}_\theta(\boldsymbol{z}_t,t)$:
\begin{equation}
\hat{\boldsymbol{\epsilon}}_\theta(\boldsymbol{z}_t,\boldsymbol{c},t)=w\boldsymbol{\epsilon}_\theta(\boldsymbol{z}_t,\boldsymbol{c},t)+(1-w)\boldsymbol{\epsilon}_\theta(\boldsymbol{z}_t,t),
\label{eq:free}
\end{equation}
where $\boldsymbol{z_{t}}={\mathcal{E}}(\boldsymbol{x_{t}})$. $w$ often named guidance scale or guidance weight, is a scalar value that adjusts the alignment with condition $\boldsymbol{c}$.

\subsection{Overview}
As shown in the Figs.~\ref{fig:framework}, our BrainDreamer adopts a two-stage pipeline, which is effective and robust. First, we leverage the pre-trained CLIP image encoder and text encoder to assist us in training the EEG encoder. We design a mask-based triple contrastive learning strategy (see Sec.~\ref{sec:cl}) to map EEG embeddings into the CLIP embedding space. Subsequently, we construct an EEG adapter(see Sec.~\ref{sec:eeg}), consisting of an EEG encoder and a feature projection module, where the parameter weights of the EEG encoder remain fixed. The EEG Adapter injects EEG embeddings into the pre-trained Stable Diffusion model in a FiLM manner, which offers a lower computational overhead compared to commonly-used cross-attention method. It is worth noting that textual descriptions are not mandatory during the generation. However, we encourage inputting a small amount of abstract textual descriptions to assist the model in better generating the corresponding images. This is because the semantic information conveyed by brain signals often represents an object alone, lacking background, color, or spatial information~\cite{bai2023dreamdiffusion}. Finally, both text embeddings extracted by CLIP text encoder and EEG embeddings are fed into the pre-trained Stable Diffusion to accomplish controlled high-quality image generation.

\subsection{Mask-based Triple Contrastive Learning}
\label{sec:cl}
Our method is primarily based on a pre-trained Stable Diffusion model, which primarily handles latent features in the CLIP embedding space. Therefore, we aim to map EEG embeddings to the CLIP embedding space, enabling pre-trained Stable Diffusion to generate higher-quality images based on EEG embeddings. Previous works~\cite{bai2023dreamdiffusion,singh2024learning} have attempted to align the EEG embeddings with the CLIP image embeddings, and then input the EEG embeddings to the generative model for image generation. We believe that such image-EEG alignment is incomplete due to the following reasons: \textbf{1)} Most generative models are text-to-image models (\eg, Imagen and Stable Diffusion), which are more sensitive to text embeddings. Despite CLIP's impressive image-text representation capabilities, there still exists a certain degree of domain gap between image and text modalities. \textbf{2)} Textual information offers greater flexibility and controllability, which is why text embeddings are often considered as ``ground truth'' in most multi-modal works~\cite{zhao2023chatbridge,wei2023diffusion}. Therefore, we also incorporate textual information to map EEG embeddings to the CLIP embedding space in addition to image information. We design a mask-based triple contrastive learning strategy for embedding alignment. Given the EEG encoder $\boldsymbol{F}_{E}$, the frozen CLIP image encoder $\boldsymbol{F}_{I}$ and the frozen CLIP text encoder $\boldsymbol{F}_{T}$, along with a sampled batch of triplets ${(E_{i},I_{i},{T}_{i})}$ for the EEG signals, its corresponding observed image, and the associated text, the contrastive loss for alignment is formulated as:

\begin{figure*}[t!]
    \centering
    \includegraphics[width=.9\textwidth]{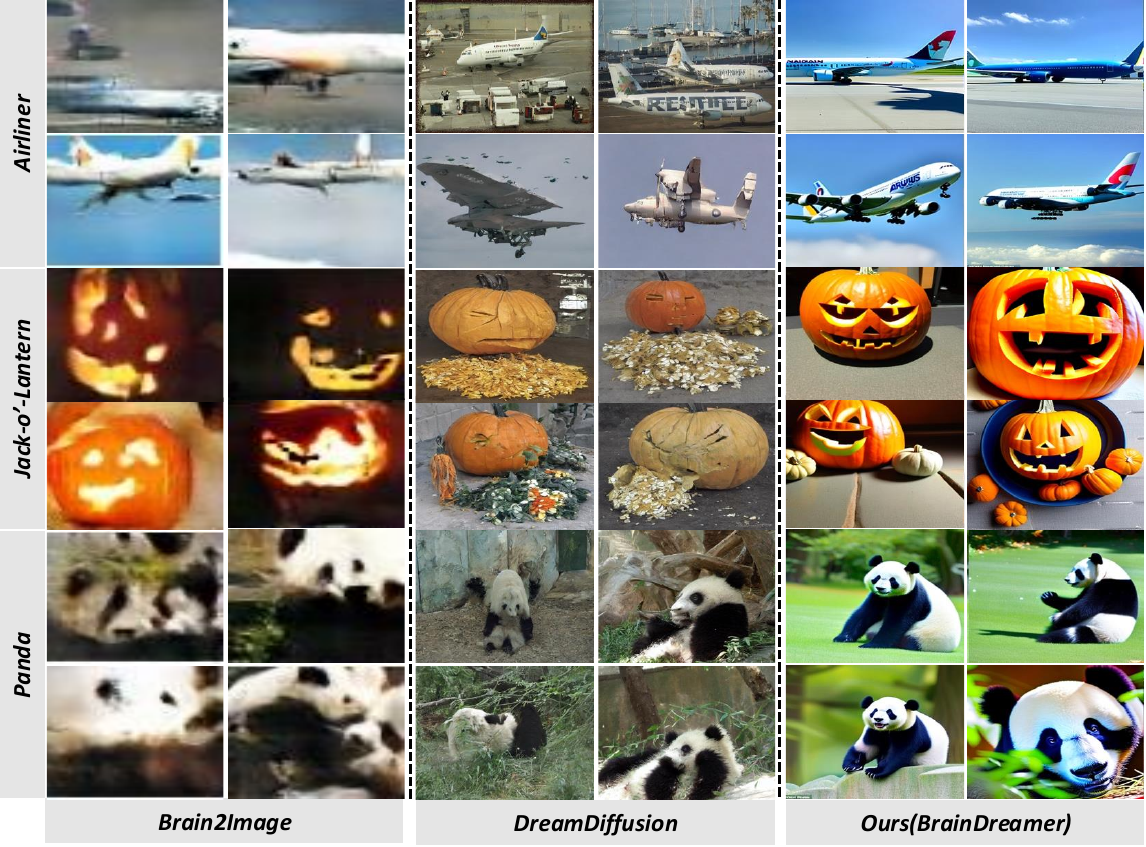}
    \vspace{-10pt}
    \caption{Qualitative comparison with Brain2Image~\cite{kavasidis2017brain2image}, DreamDiffusion~\cite{bai2023dreamdiffusion} for EGG-to-Image generation (\textbf{Setting 1}).
    }
    \label{fig:compare_with_previous}
    \vspace{-10pt}
\end{figure*}

\begin{small}
\begin{equation}
    \begin{aligned}
    \mathcal{L}=-\frac{1}{6B}\sum_{i=1}^{B}\left(\underbrace{\log\frac{\exp(f_i^E\cdot f_i^I/\tau)}{\sum_j\exp(f_i^E\cdot f_j^I/\tau)}+\log\frac{\exp(f_i^I\cdot f_i^E/\tau)}{\sum_j\exp(f_i^I\cdot f_j^E/\tau)}}_{\mathcal{L}_{EI}:\text{EEG-Image contrastive}}\right. 
    \\
    +\underbrace{\log\frac{\exp(f_i^E\cdot f_i^T/\tau)}{\sum_j\exp(f_i^E\cdot f_j^T/\tau)}+\log\frac{\exp(f_i^T\cdot f_i^E/\tau)}{\sum_j\exp(f_i^T\cdot f_j^E/\tau)}}_{\mathcal{L}_{ET}:\text{EEG-Text contrastive}}
    \\
    \left.+\underbrace{\log\frac{\exp(f_i^I\cdot f_i^T/\tau)}{\sum_j\exp(f_i^I\cdot f_j^T/\tau)}+\log\frac{\exp(f_i^T\cdot f_i^I/\tau)}{\sum_j\exp(f_i^T\cdot f_j^I/\tau)}}_{\mathcal{L}_{IT}:\text{Image-Text contrastive}}\right ),
    \end{aligned}
    \label{eq:loss}
\end{equation}
\end{small}
where $B$ is the number of shapes in a batch; $\tau$ is a learnable temperature; $f_{i}^{E}=\operatorname{Norm}(F_{E}(E_{i}))$, $f_{i}^{I}=\operatorname{Norm}(F_{I}(I_{i}))$, $f_{i}^{T}=\operatorname{Norm}(F_{T}(T_{i}))$ and $\operatorname{Norm}$ is normalization. In addition, we employ masked modeling on the image and EEG data. This not only enhances the robustness of the features but also reduces training costs. The objective of training the EEG encoder is to minimize $\mathcal{L}$.

\subsection{EEG Adapter}
\label{sec:eeg}
EEG adapter is designed to enable the pre-trained Stable Diffusion model to generate images with EEG signals. As shown in Fig.~\ref{fig:framework}, the EEG adapter consists of a frozen EEG encoder and a feature projection module. Previous methods~\cite{bai2023dreamdiffusion} simply feed EEG embeddings into the frozen cross-attention layers of Stable Diffusion to generate corresponding images, overlooking the domain gap between EEG embeddings and text embeddings, despite prior efforts to align the embeddings. To address this issue, we introduce a feature projection module to reduce the domain gap between the embeddings. Furthermore, we employ the FiLM mechanism to inject embeddings into models instead of common cross-attention methods. Previous works, \eg,~\cite{dit} have demonstrated that FiLM is more suitable for incorporating category information (\eg, EEG signals), while the cross-attention method is more applicable to sequential information (\eg, text description). In addition, compared to the additional 15\% Gflops overhead introduced by cross-attention, the computational cost associated with FiLM is almost negligible~\cite{dit}.
Given the query features $\mathbf{Z}$ and the EEG embeddings $\boldsymbol{c}_{e}$, the output of EEG adapter is computed as follows:
\begin{equation}
\begin{aligned}
    \boldsymbol{\alpha}, \boldsymbol{\beta} = \operatorname{FP}(\boldsymbol{c}_{e}),\\
    \mathbf{Z}^{\prime\prime}= \mathbf{Z}\odot\boldsymbol{\alpha} + \boldsymbol{\beta},
\end{aligned}
\end{equation}
where $\odot$ denotes element-wise multiplication and the $\operatorname{FP}(\cdot)$ is the feature projection module which consists of two linear layers, a normalization layer, and an activation function (see Fig.~\ref{fig:framework}).

\subsection{Image Generation based on EEG and Texts}
Although textual descriptions are not mandatory, BrainDreamer encourages the input of textual descriptions to assist in generating desired images. Given the query features $\mathbf{Z}$ and the text embeddings $\boldsymbol{c}_{t}$, the output of cross-attention $\mathbf{Z}^{\prime}$ can be defined as follows:
\begin{equation}
    \mathbf{Z}^{\prime}=\text{Attention}(\mathbf{Q},\mathbf{K},\mathbf{V})=\text{Softmax}(\frac{\mathbf{Q}\mathbf{K}^{\top}}{\sqrt{d}})\mathbf{V},
\end{equation}
where $\mathbf{Q}=\mathbf{Z}\mathbf{W}_q$, $\mathbf{K}=\boldsymbol{c}_t\mathbf{W}_k$, $\mathbf{V}=\boldsymbol{c}_t\mathbf{W}_v$ are the query, key, and values matrices of the attention operation respectively, and $\mathbf{W}_q$, $\mathbf{W}_k$, $\mathbf{W}_v$ are the weight matrices of the linear projection layers. Then we simply add the output of EEG adapter to the output of text cross-attention.  This process is formulated as follows:
\begin{equation}
\mathbf{Z}^{new}=\text{Softmax}(\frac{\mathbf{Q}\mathbf{K}^{\top}}{\sqrt{d}})\mathbf{V} + \lambda(\mathbf{Z}\odot\boldsymbol{\alpha} + \boldsymbol{\beta}).
\end{equation}
where $\lambda$ is weight factor, and the model becomes the original text-to-image diffusion model if $\lambda=0$. Hence, the training objective Eq.~\ref{eq:L_dm} is updated to:
\begin{equation}
\mathcal{L}=\mathbb{E}_{z,\epsilon,c_t,c_e,t}[\|\boldsymbol{\epsilon}_{t}-\boldsymbol{\epsilon}_\theta(\boldsymbol{z}_t,\boldsymbol{c}_t,\boldsymbol{c}_e,t)\|^2].
\end{equation}
And the Eq.~\ref{eq:free} is updated to:
\begin{equation}\hat{\boldsymbol{\epsilon}}_\theta(\boldsymbol{z}_t,\boldsymbol{c}_{t},\boldsymbol{c}_{e},t)=w\boldsymbol{\epsilon}_\theta(\boldsymbol{z}_t,\boldsymbol{c}_{t},\boldsymbol{c}_{e},t)+(1-w)\boldsymbol{\epsilon}_\theta(\boldsymbol{z}_t,t).
\end{equation}

\section{Experiments}
\label{sec:exp}
\subsection{Implementation Details and Benchmark Datasets}
\noindent \textbf{Implementation Details.}
We align with the optimal settings established by~\cite{bai2023dreamdiffusion}, and adjust our parameters: EEG time series length as 512, embedding dimension as 1280, channels as 128.

In the first stage of our fine-tuning of the EEG encoder, only the parameters of the EEG encoder are updated, and the text encoder and image encoder are frozen at this step. The joint fine-tuning of the EEG encoder made it compatible with the vanilla CLIP model. The mask ratio for image and EEG data is set to 0.5. For CLIP, we use the ViT-L/14 model to extract feature embeddings. The architecture of EEG encoder is the same as ViT-Large~\cite{vit}. Regarding our generative model architecture, we utilize version 1.5 of the Stable Diffusion model. There are 16 cross-attention layers in this model, and we add a new EEG Adapter for each of these layers. 

In the second stage of training the EEG Adapter, we use the AdamW optimizer with a fixed learning rate of 0.0001 and weight decay of 0.01. To enable classifier-free guidance, we drop EEG embeddings and text embeddings individually by using a probability of 0.05 and drop EEG embeddings and text embeddings simultaneously by using a probability of 0.05. In the inference stage, we adopt DDIM~\cite{ddim} sampler with 50 steps and set the guidance scale $w$ to 7.5. When only using the EEG prompt, we set the text prompt to empty and $\lambda$ = 1.0.

\noindent \textbf{Benchmark Datasets.}
In our research, we utilize the EEG-image dataset~\cite{kavasidis2017brain2image} to verify the effectiveness of our method. EEG-image contains EEG signals paired with 2,000 images from ImageNet, which includes 11,466 EEG sequences captured through 128-channel electrodes. These EEG recordings corresponded with 40 unique image classes, with each class represented by 50 distinct images sourced from the ImageNet dataset. The EEG recordings were obtained from 6 subjects, and our training and testing were conducted on the same subject, specifically, Subject 4.

\begin{figure}[t!]
    \centering
    \includegraphics[width=\linewidth]{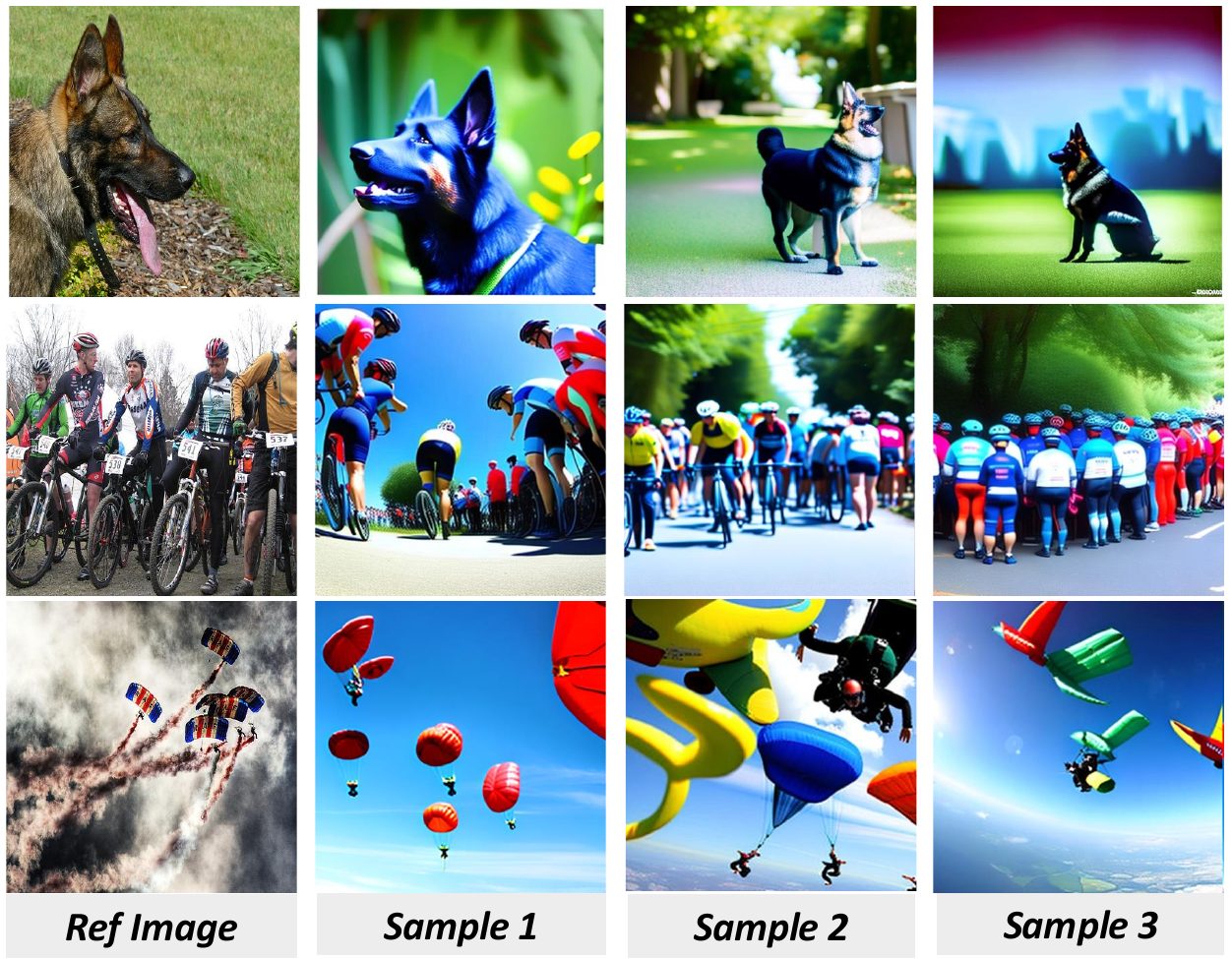}
    \vspace{-10pt}
    \caption{Some results of directly generating images from EEG signals using BrainDreamer (\textbf{Setting 1}). The images on the left depict paired image data, while the three images on the right represent the sampling results.}
    \vspace{-10pt}
    \label{fig:res_no}
\end{figure}

\begin{figure}[t!]
    \centering
    \includegraphics[width=\linewidth]{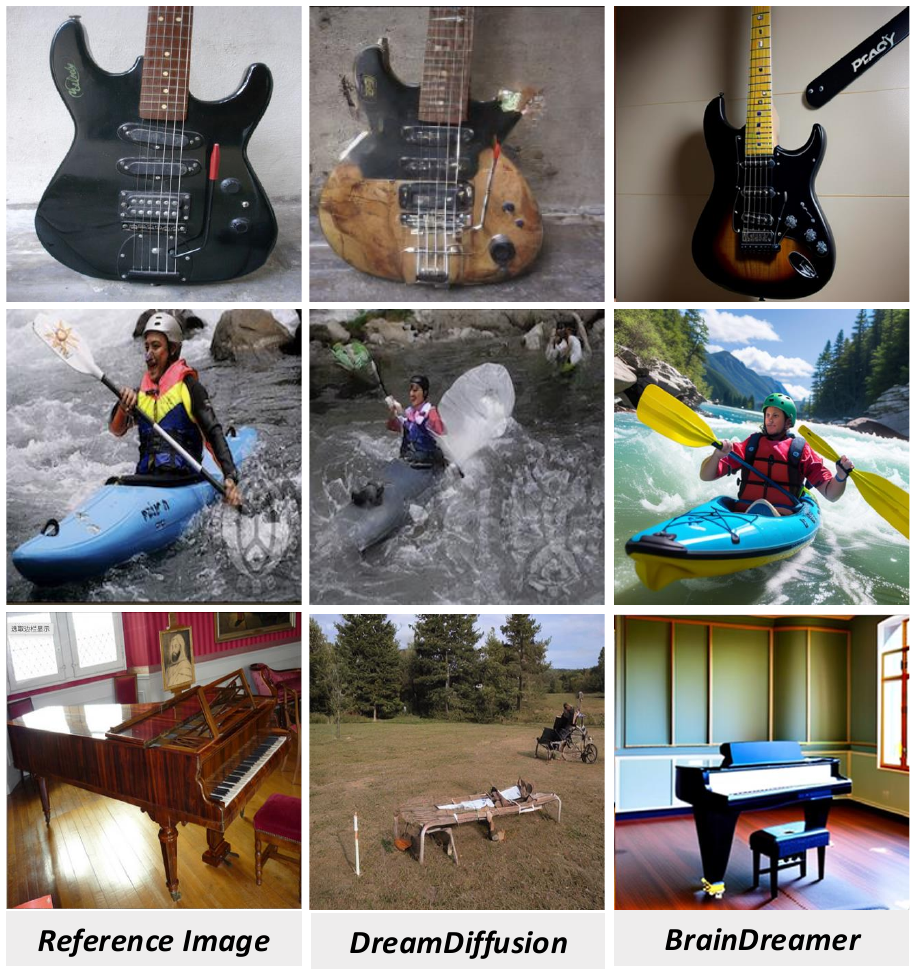}
    \vspace{-15pt}
    \caption{Qualitative comparison of EGG-to-Image generation (\textbf{Setting 1}) between DreamDiffusion~\cite{bai2023dreamdiffusion} and ours with the reference to the GT on the EEG-image dataset.}
    \label{fig:compare_with_dreamdiffusion}
\end{figure}

\begin{figure*}[t!]
    \centering
    \includegraphics[width=.85\linewidth]{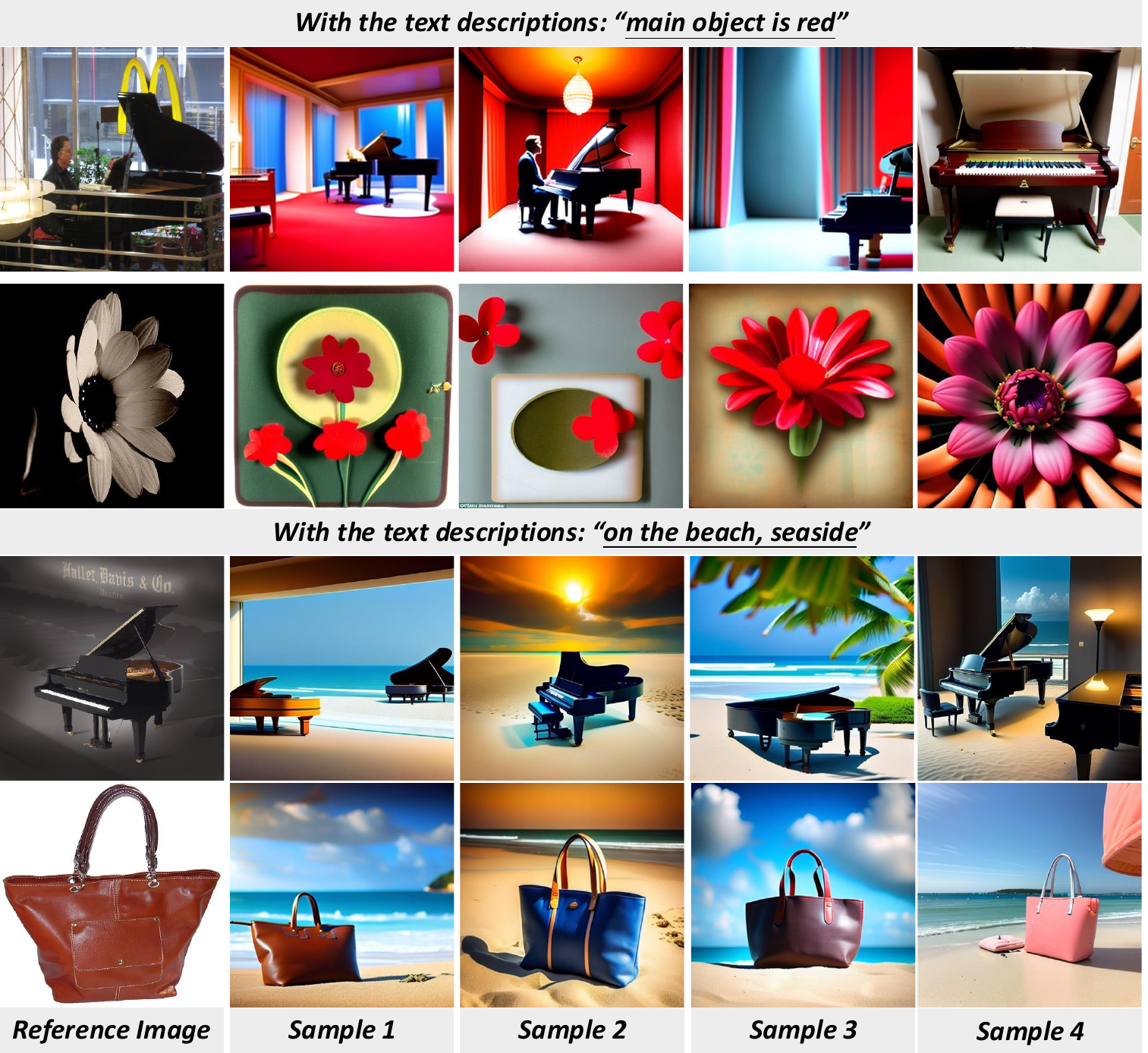}
     \vspace{-7pt}
    \caption{Visual results of generating images from EEG signals with text descriptions using BrainDreamer (\textbf{Setting 2}). It can be seen that our method effectively combines textual descriptions and creates reasoning-coherent images from EEG signals.}
    \label{fig:res}
\end{figure*}

\subsection{Experimental results}
\noindent \textbf{Experimental settings.} Our experiment is conducted in two settings.  1) \textbf{EEG-to-Image Generation}. It indicates generating images from only EGG input. 2) \textbf{(EEG+Text)-to-Image Generation}. In this setting, we also provide textual descriptions related to color and background to guide EEG-to-image generation.
EEG Classification Accuracy (ECA) and CLIP Similarity (CS) are employed for quantitative comparisons in our experiments. ECA reflects the classification performance of the generated images, providing a coarse-grained assessment of semantic consistency between the generated images and GT. We employ a pre-trained ImageNet1K~\cite{vit} classifier to calculate the top-1 classification accuracy and evaluate the semantic accuracy of the generated images. CS evaluates the semantic similarity between the generated images and ground truth by utilizing the pre-trained CLIP ViT-B/32 image encoder to extract image embeddings and compute their cosine similarity.

\noindent \textbf{EEG-to-Image Generation (Setting 1).}
Since reference implementations are not available for most of the related works, we only compare our approach with the recent work, DreamDiffusion. Referring to the settings of DreamDiffusion, we report the quantitative comparison results in Tab.~\ref{tab:res}. It can be observed that our method outperforms existing approaches in terms of semantic consistency in the generated images. We also conduct visual comparisons, and the results are displayed in Figs.~\ref{fig:compare_with_previous} and~\ref{fig:compare_with_dreamdiffusion}. We provide more visual results of our BrainDreamer in Fig.~\ref{fig:res_no}. Our method outperforms existing approaches in terms of image generation quality.

\begin{table}[t]
\centering
\caption{Quantitative comparative experiment of reconstruction results. We compare our BrainDreamer with DreamDiffusion.}
\vspace{-5pt}
\label{tab:res}
\begin{tabularx}{\linewidth}{>{\centering\arraybackslash}p{0.4\linewidth}|>{\centering\arraybackslash}X>{\centering\arraybackslash}X}
\toprule[0.2mm]
                         & ECA  & CS \\ \hline
DreamDiffusion~\cite{bai2023dreamdiffusion}       & 29.9 &   55.8\\
Ours (on Brain2Image) &   \textbf{44.5}  &  \textbf{68.1}  \\ 
\toprule[0.2mm]
\end{tabularx}
\end{table}
\begin{table}[t!]
    \centering
    \caption{Ablation studies of EEG adapter. ``Param'' refers to the additional number of parameters introduced by a certain method.}
    \vspace{-5pt}
    \label{tab:eeg}
    \begin{tabularx}{\linewidth}{>{\centering\arraybackslash}p{0.3\linewidth}|>{\centering\arraybackslash}X>{\centering\arraybackslash}X|>{\centering\arraybackslash}X}
    \toprule[0.2mm]
                          & ECA (\%) & CS (\%) & Param (M)\\ \hline
    Direct injection~\cite{bai2023dreamdiffusion}      & 28.7   &  22.5     & 0\\ 
    Cross-attention       & 37.8   &  58.9     & 22 \\ 
    Ours                  & \textbf{44.5}  &  \textbf{68.1} & 5 \\ 
    \toprule[0.2mm]
    \end{tabularx}
    \vspace{-5pt}
\end{table}
\begin{table}[t]
    \centering
    \caption{Ablation studies of mask-based triple contrastive learning.}
    \vspace{-5pt}
    \label{tab:cl}
    \begin{tabularx}{\linewidth}{>{\centering\arraybackslash}p{0.4\linewidth}|>{\centering\arraybackslash}X>{\centering\arraybackslash}X}
    \toprule[0.2mm]
                         & ECA (\%) & CS (\%) \\ \hline
    w/o masked modeling  &   42.6  & 61.9 \\
    w/o text supervision &   31.7  & 56.3      \\
    Ours                 &   \textbf{44.5}  &  \textbf{68.1}  \\ 
    \toprule[0.2mm]
    \end{tabularx}
\end{table}
\begin{table}[t!]
    \centering
    \caption{Ablation studies of mask ratios.}
    \vspace{-5pt}
    \label{tab:mask}
    \begin{tabularx}{\linewidth}{>{\centering\arraybackslash}p{0.4\linewidth}|>{\centering\arraybackslash}X>{\centering\arraybackslash}X}
    \toprule[0.2mm]
    Mask ratios & ECA (\%) & CS (\%) \\ \hline
    0.3         &  42.1   &    62.9 \\
    0.5         &  \textbf{44.5}  &  \textbf{68.1}\\
    0.75        &  41.5   &    63.2\\
    \toprule[0.2mm]
    \end{tabularx}
    \vspace{-10pt}
\end{table}
\noindent \textbf{(EEG + Text)-to-Image Generation (Setting 2).}
We are the first to attempt to use textual descriptions as guidance for generating images from EEG brain signals, resulting in a method that aligns better with human visual perception. We provide some visual examples, which can be seen in Figs.~\ref{fig:res} and ~\ref{fig:res_text}. As can be observed, our method effectively combines textual descriptions to generate high-quality reasoning-coherent images from EEG brain signals.

\subsection{Ablation Study}
To assess the impact of individual components of our method, we conducted a series of ablation experiments. All variants are trained using the same settings.

\noindent \textbf{Effectiveness of EEG adapter.}
To validate the effectiveness of the EEG adapter, we remove the feature projection module and inject the EEG signals into the pre-trained Stable Diffusion model following the methodology of DreamDiffusion. In addition, we also attempt to compare with standard cross-attention. The results are shown in Fig.~\ref{tab:eeg}. Clearly, our method has achieved a better trade-off between performance and parameters.

\noindent \textbf{Effectiveness of mask-based triple contrastive learning.}
We verify the effectiveness of the proposed mask-based triple contrastive learning strategy in Fig.~\ref{tab:cl}. The results show that both removing the masked modeling and abandoning text supervision have had a detrimental impact on our results. 

\noindent \textbf{Mask ratios.}
We investigate to best mask ratio for mask-based triple contrastive learning. As shown in Fig.~\ref{tab:mask}, both excessively high and low mask ratios can have detrimental effects on the model. Excessive mask ratios may result in the loss of valuable information, while too low mask ratios may fail to reduce memory overhead, resulting in a smaller batch size for contrastive learning. 

\subsection{Disccusion}
\subsubsection{Analysis about the Necessity of Textual Guidance Interaction}
\begin{figure}[t!]
    \centering
     \vspace{-10pt}
     \includegraphics[width=\linewidth]{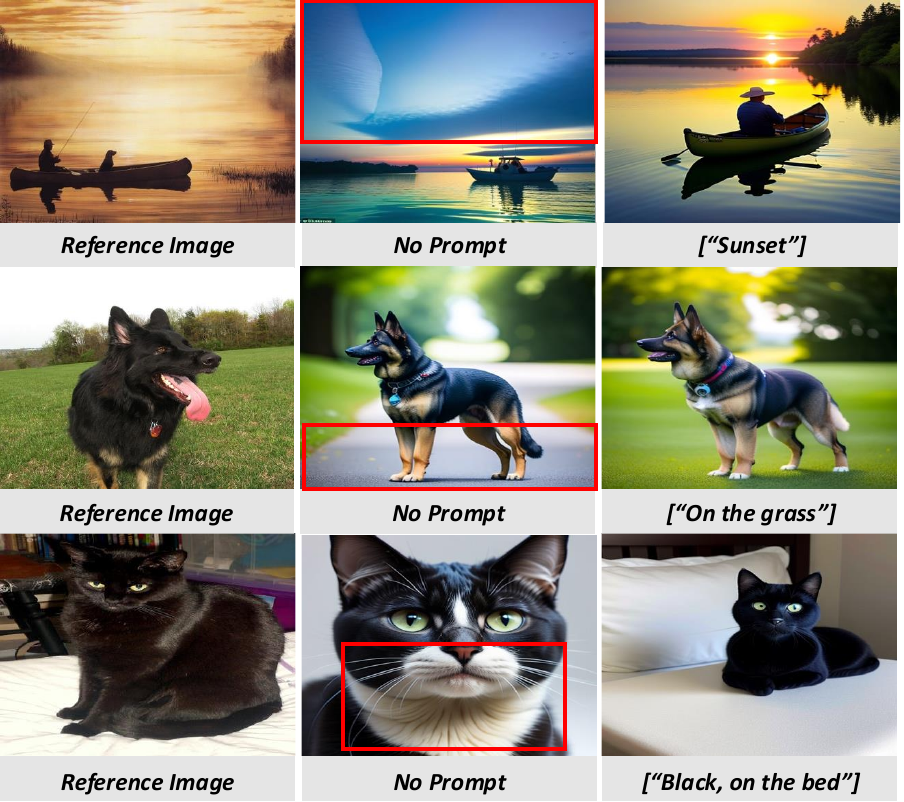}
    \caption{Examples of reasoning-coherent image generation with the textual guided interaction.}
    \label{fig:reasoning}
\end{figure}

EEG signals provide coarse-grained information about the main object, while text offers \textit{background} information that EEG cannot provide; combining EEG with text for image generation is reasoning-coherent because they are complementary.  The textual descriptions are only about the content of the background or the object's fine-grained features like color, without providing information about the main object, which is only provided by EEG signals. As shown in the first two columns of Fig.~\ref{fig:reasoning}, images directly generated from EEG, although they reflect the correct primary object information that it is a boat, dog, and cat, respectively, the colors and the background information are incorrect. If one wishes to generate a complete and correct image through EEG signals, the complete user interaction logic is that the user completes the imagination, the helmet collects the EEG signals, and then the user describes the finer-grained features, and finally, the computer completes the generation of an image that is correct in the main body and rich in details as imagined by the user.

This process of combining EEG and text for image generation is analogous to the process of creating a suspect sketch, where the sketch artist begins with a coarse-grained description of the suspect based on the eyewitness's overall impression, such as gender, race, and general body shape. This is similar to how EEG signals provide a rough representation of the main object in the user's imagination. As the sketch progresses, the artist collects more fine-grained details, like specific facial features, hairstyles, and clothing, to refine the image. Similarly, in the EEG-text fusion approach, the EEG provides the foundational image information about the main object, while text descriptions add the finer details and background information that the EEG cannot capture. Both processes rely on the complementary strengths of two sources of information to create a more accurate and complete representation. Just as the suspect sketch becomes more precise through iterative detailing based on eyewitness feedback, the EEG-text image generation process also follows a logic that mirrors human cognition: the user imagines the main object, the EEG captures the core features, and text inputs fill in the missing finer details. This analogy illustrates the reasonableness of the EEG-text approach, highlighting how the combination of these two modalities creates a coherent, detailed, and contextually rich image that faithfully represents the user's mental picture.

\subsubsection{Explanation about Reasoning-coherent Interaction of EEG-Image Generation}
EEG signals provide coarse-grained information about the main object, \eg, `Cat', while text offers details and background information, \eg, `Black, on the bed', that EEG cannot provide. Therefore, the generated image could be a black cat on the bed. Such an image generation from EEG and language is regarded as reasoning-coherent because it mimics human imagination. 

This method represents an innovative and novel form of human-computer interaction, where the brain's neural activity, captured through EEG signals, directly interfaces with computational models to create visual content. By integrating EEG signals with textual descriptions, this approach leverages the strengths of both modalities: the intuitive, subconscious processing of the brain for identifying main objects, and the deliberate, conscious articulation provided by language for specifying details and context. This dual-modality interaction not only allows for a more natural and intuitive user experience but also opens up new possibilities for users to express their imagination and creativity. Unlike traditional image generation methods that rely solely on explicit text input or pre-defined templates, this EEG-text fusion approach dynamically translates human thought patterns into rich visual representations, fostering a deeper connection between human cognitive processes and machine-generated outputs. By aligning closely with how humans naturally envision and describe their surroundings, this technique enhances user engagement and could pave the way for more immersive applications in areas, such as virtual reality, personalized content creation, and cognitive neuroscience research.

\subsection{Evaluations on Real-world EEG Data}
\begin{figure}[t!]
    \centering
    \includegraphics[width=\linewidth]{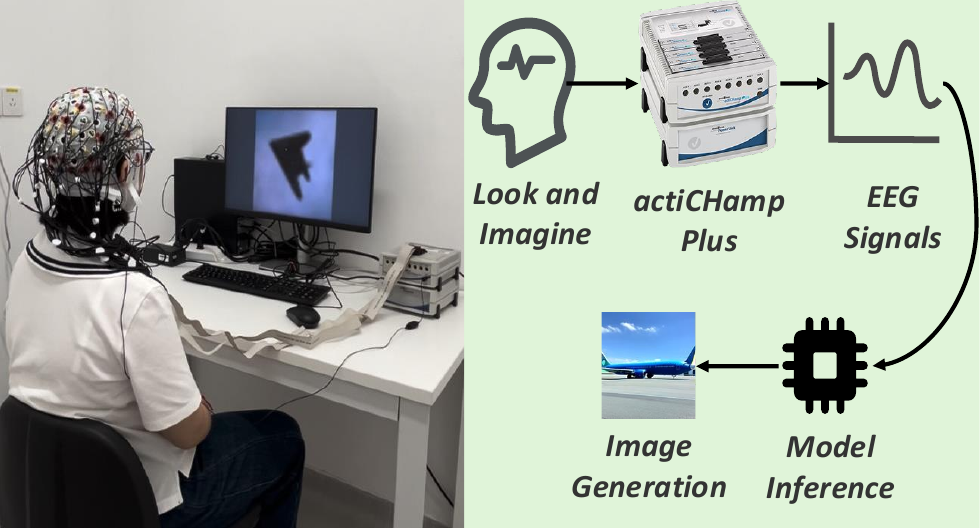}
    \vspace{-15pt}
    \caption{The participant was equipped with actiCAP and actiCHamp Plus to collect her EEG brain signals.}
    \label{fig:user_study}
    \vspace{-15pt}
\end{figure}

To evaluate the effectiveness and user experience of BrainDreamer, we conducted a user study with several participants. The EEG signal collection and study procedures were approved by the ethical committee. After collecting EEG signals, we fine-tuned the EEG encoder using real-world data. The study’s objective was to assess the feasibility of generating images from EEG signals and the effectiveness of our proposed method. Specifically, it aimed to assess how well users could interpret these generated images with and without accompanying textual guidance and to measure the perceived quality and coherence of the images produced.

\noindent \textbf{Participants and Devices.}
Due to the difficulty of collecting EEG signals, we recruited 2 university students (1 male and 1 female) as participants. 
We collect EEG signals when participants look at the image stimuli.
To collect brain responses, participants were wearing electrodes and caps (actiCAP\footnote{\url{https://www.brainproducts.com/solutions/acticap/}}), which were connected to amplifiers (actiCHamp Plus\footnote{\url{https://brainvision.com/products/actichamp-plus/}}), and the amplified signals were finally recorded by BrainVision Recorder\footnote{\url{https://www.brainproducts.com/solutions/recorder/}}. We organized three categories of images from ImageNet, which are fish, butterfly, and plane, and selected 20 images for each category. Participants were asked to continuously look at the images displayed on the screen and imagine the corresponding objects. Our image display process follows the Rapid Serial Visual Presentation (RSVP) paradigm\cite{THINGSEEG1,intraub1981rapid,keysers2001speed}.
\begin{figure}[t!]
    \centering
    \includegraphics[width=\linewidth]{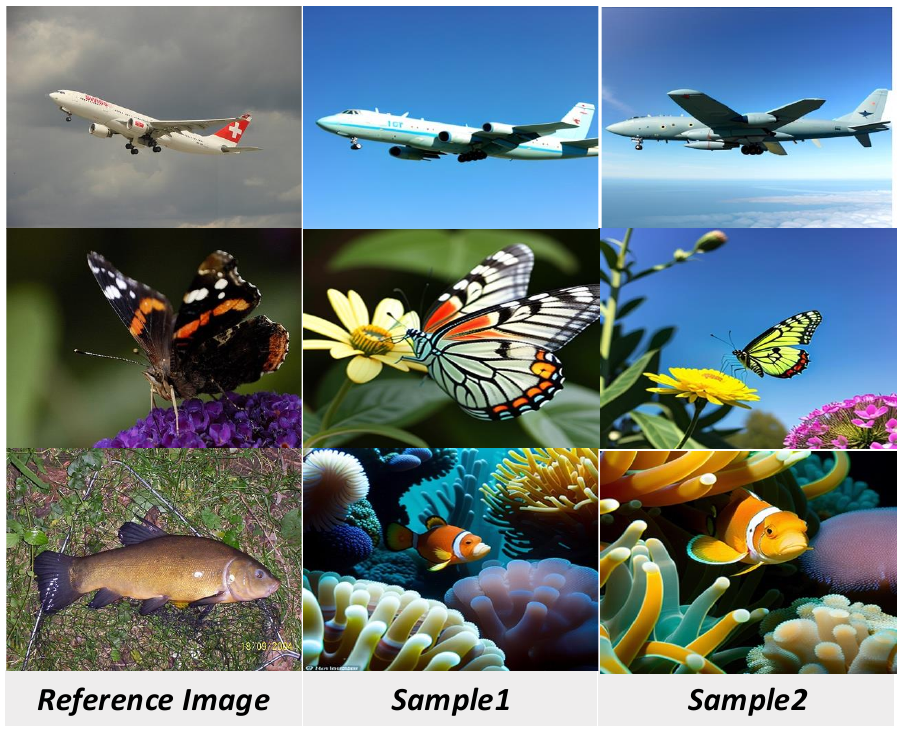}
    \vspace{-15pt}
    \caption{Generation results of real-world EEG data.}
    \label{fig:realworld}
    \vspace{-15pt}
\end{figure}

\noindent \textbf{Task Design.}
The user study consisted of two main phases:
\begin{enumerate}
    \item \textbf{EEG-to-Image Generation (Without Textual Guidance)}: Participants were asked to look and imagine specific scenes or objects while their EEG signals were recorded. BrainDreamer then generated images based solely on these EEG signals.
    \item \textbf{EEG + Text-to-Image Generation (With Textual Guidance)}: In this phase, participants provide a text prompt describing some lost fine-grained information based on phase 1 generation results. The EEG signals were then used in conjunction with the text prompt to generate an image.
\end{enumerate}
\noindent \textbf{Comparative Analysis and User Feedback.}
A comparative analysis of the two phases showed that images generated with the aid of textual guidance were generally rated higher on scales of clarity, detail, and overall satisfaction. On average, participants rated the EEG-only generated images with a mean score of 3.2 out of 5 for interpretability, whereas the images generated with EEG and textual guidance received a mean score of 4.5. Participants noted that the EEG-only images often appeared abstract or blurred, which made them challenging to interpret without additional context. In contrast, the guided images were described as more vivid and clear, with elements that were easier to recognize and align with the participants’ mental imagery.

\noindent \textbf{Perceived Quality and Coherence.} 
The perceived quality and coherence of the images were also assessed. Images generated from EEG signals alone were often perceived as less coherent, with participants expressing difficulty in making sense of the abstract shapes and forms. However, the addition of textual guidance appeared to bridge the gap between abstract EEG representations and more concrete visual outcomes. This suggests that while EEG signals contain valuable information about a participant’s mental imagery, textual cues are crucial for refining and contextualizing this information to produce images that are not only coherent but also more closely aligned with the participant’s intended visualization.

\section{Conclusion}
\label{sec:conclusion}
In this paper, we proposed a novel end-to-end language-guided generative framework BrainDreamer that can mimic human reasoning and generate high-quality images from EEG brain signals. We designed a novel mask-based triple contrastive learning strategy to effectively align EEG, text, and image embeddings. Moreover, we have designed an EEG adapter to inject EEG embeddings into a pre-trained Stable Diffusion model, enabling us to achieve reasoning-coherent image generation. In the generation stage, we can provide abstract textual descriptions (\eg, background information) to assist in the image generation, ensuring that the results align more closely with our expectations. Our BrainDreamer is more aligned with human visual perception, and the generation quality and quantitative performance are better than existing methods. We also conducted a real-world user study to confirm the effectiveness of BrainDreamer and show its practical potential for applications like personalized content creation in virtual reality. Its ability to generate reasoning-coherent and controllable images from brain signals takes a step forward in human-computer interaction, bridging the gap between human imagination and machine-generated visual content. Future work will focus on enhancing the model’s robustness and generalization to variations in EEG signals, while also expanding the framework’s potential applications.

\begin{figure}[t!]
    \centering
    \includegraphics[width=.75\linewidth]{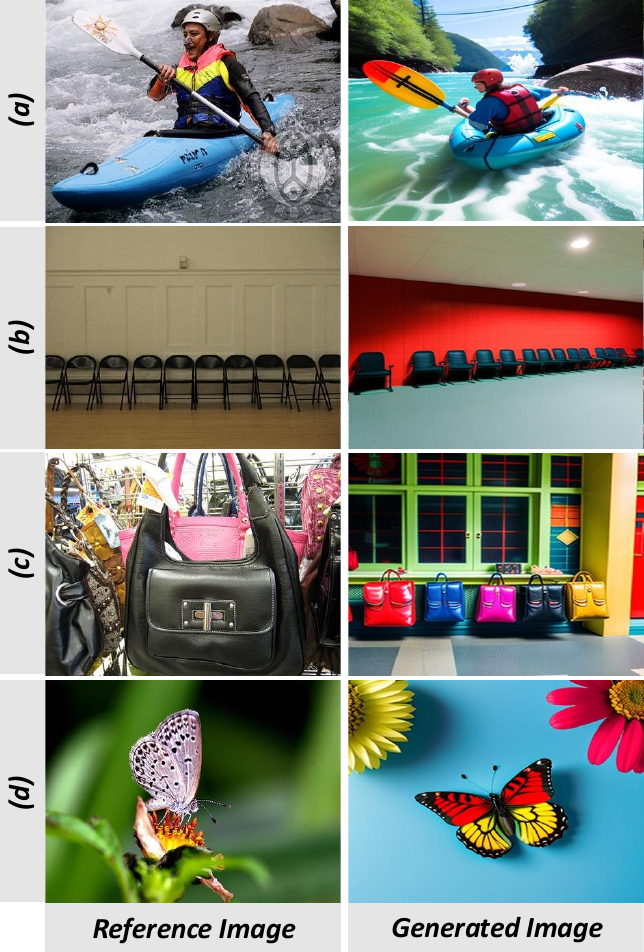}
    \vspace{-12pt}
    \caption{Some unsatisfactory results of our BrainDreamer. In these cases, we use the \textsf{[``main object is red.'']} as the text description. As you can see, BrainDreamer exhibits instance mismatch and insufficient coloring issues on these generated images.}
    \label{fig:limit}
    \vspace{-14pt}
\end{figure}

\noindent \textbf{Limitations.} In Fig.~\ref{fig:limit}, we show some unsatisfactory results of our BrainDreamer. Although these results exhibit a high level of coarse-grained matching with the ground truth and indeed interact with our textual descriptions, they are still imperfect. In cases (a) and (b), our BrainDreamer fails to recognize the main subjects in the images, resulting in instance mismatch. Particularly in (b), the wall is incorrectly colored red instead of the chair. In cases (c) and (d), BrainDreamer exhibits insufficient coloring, particularly in (d), where only half of the butterfly is red. Such images are highly unrealistic. Combining EEG signals with textual descriptions or other supplementary information to achieve instance-level image generation will be the focus of our future work.

\clearpage
\bibliographystyle{abbrv-doi}

\bibliography{main}
\end{document}